\PassOptionsToPackage{table}{xcolor}
\documentclass[10pt,twocolumn,letterpaper]{article}
\usepackage[pagenumbers]{cvpr}

\usepackage{multirow}
\usepackage{float}
\usepackage{array}
\usepackage{tabularx}
\usepackage{wrapfig}
\usepackage{xurl}
\usepackage{needspace}

\providecommand{\papertext}[2]{#1}

\definecolor{cvprblue}{rgb}{0.21,0.49,0.74}
\usepackage[pagebackref,breaklinks,colorlinks,allcolors=cvprblue]{hyperref}

\title{ADClaw: A Programmable Visual Agent for Industrial Anomaly \\ Detection and Understanding}
\author{%
    Xinyi Wu\textsuperscript{1,\ensuremath{\dagger},\ensuremath{\ddagger}}\quad
    Rongxin Gao\textsuperscript{1,\ensuremath{\dagger}}\quad
    YuHui Lin\textsuperscript{2}\quad
    Yonglin Hong\textsuperscript{1}\quad
    Chu Li\textsuperscript{3}\\[3pt]
    Wei Li\textsuperscript{1}\quad
    Dingzhao Li\textsuperscript{1}\quad
    Shijia Zhao\textsuperscript{1}\quad
    Chaopeng Li\textsuperscript{2}\quad
    Haixin Sun\textsuperscript{1,*}\\[5pt]
    {\small \textsuperscript{1}Xiamen University\quad
     \textsuperscript{2}Jimei University\quad
     \textsuperscript{3}China University of Mining and Technology}\\[2pt]
    {\small
     \textsuperscript{\ensuremath{\dagger}}Equal contribution
     \quad
     \textsuperscript{\ensuremath{\ddagger}}Project Lead
     \quad
     \textsuperscript{*}Corresponding author}%
}
\begin{document}
\maketitle

\flushbottom

\begin{abstract}

\papertext{Industrial anomaly detection (IAD), the task of identifying and localizing deviations from normal product conditions, is essential to automated quality inspection in manufacturing.
Many vision-language models for IAD rely on large domain-specific datasets for fine-tuning and make predictions in a single pass, leaving no opportunity to re-examine ambiguous regions.
Although tool-using agents can revisit these regions, predefined tool interfaces limit how they can combine intermediate results in a follow-up check.
When such checks are insufficient, models may miss subtle defects or misidentify the defect type.
To address this limitation, we introduce \textbf{ADClaw}, an inspection agent that uses executable code to test anomaly hypotheses without task-specific fine-tuning.
For each hypothesis, a \textbf{Structured Normality Prior} identifies the applicable rules and retrieves normal visual references for the inspection target.
Guided by this prior, the agent writes and executes code in a persistent \textbf{Inspection Workbench} to combine masks, anomaly maps, and measurements from earlier steps into a new test of the suspected defect.
After each check, \textbf{Hypothesis-Guided Evidence Reasoning} assesses whether the evidence supports the suspected defect and its location, using gaps or inconsistencies in that evidence to guide the next code action.
ADClaw achieves competitive results in detection and localization across 2D and 3D industrial anomaly detection benchmarks and in anomaly understanding on MMAD.}{工业异常检测（IAD）旨在识别并定位产品中偏离正常状态之处，对制造业的自动化质量检测至关重要。许多用于 IAD 的视觉语言模型依赖大规模领域专用数据进行微调，且仅凭单次观测作出预测，因而无法复查存疑区域。尽管配备工具的智能体能够复查这些区域，预定义的工具接口仍限制了它们在后续检查中组合中间结果的能力。若未能充分验证疑点，模型就可能漏检细微缺陷，或误判缺陷类型。为此，我们提出工业检测智能体 \textbf{ADClaw}，它无需针对任务进行微调，即可通过编写并执行代码来检验异常假设。针对每个假设，\textbf{结构化正常性先验}确定适用于检测目标的规则，并检索相应的正常视觉参考。在该先验的指导下，智能体在持久化的\textbf{检测工作台}中编写并执行代码，组合先前步骤得到的掩码、异常响应图和测量结果，构造针对疑似缺陷的新检查。每次检查后，\textbf{假设引导的证据推理}评估现有证据能否支持对缺陷及其位置的判断，并依据证据中的缺口或矛盾指导下一步代码动作。ADClaw 在二维和三维工业异常检测基准上的检测与定位任务中，以及在 MMAD 的异常理解任务中，均取得了具有竞争力的结果。}

\end{abstract}

\section{\papertext{Introduction}{引言}}

\begin{figure*}[t]
    \centering
    \includegraphics[width=\textwidth]{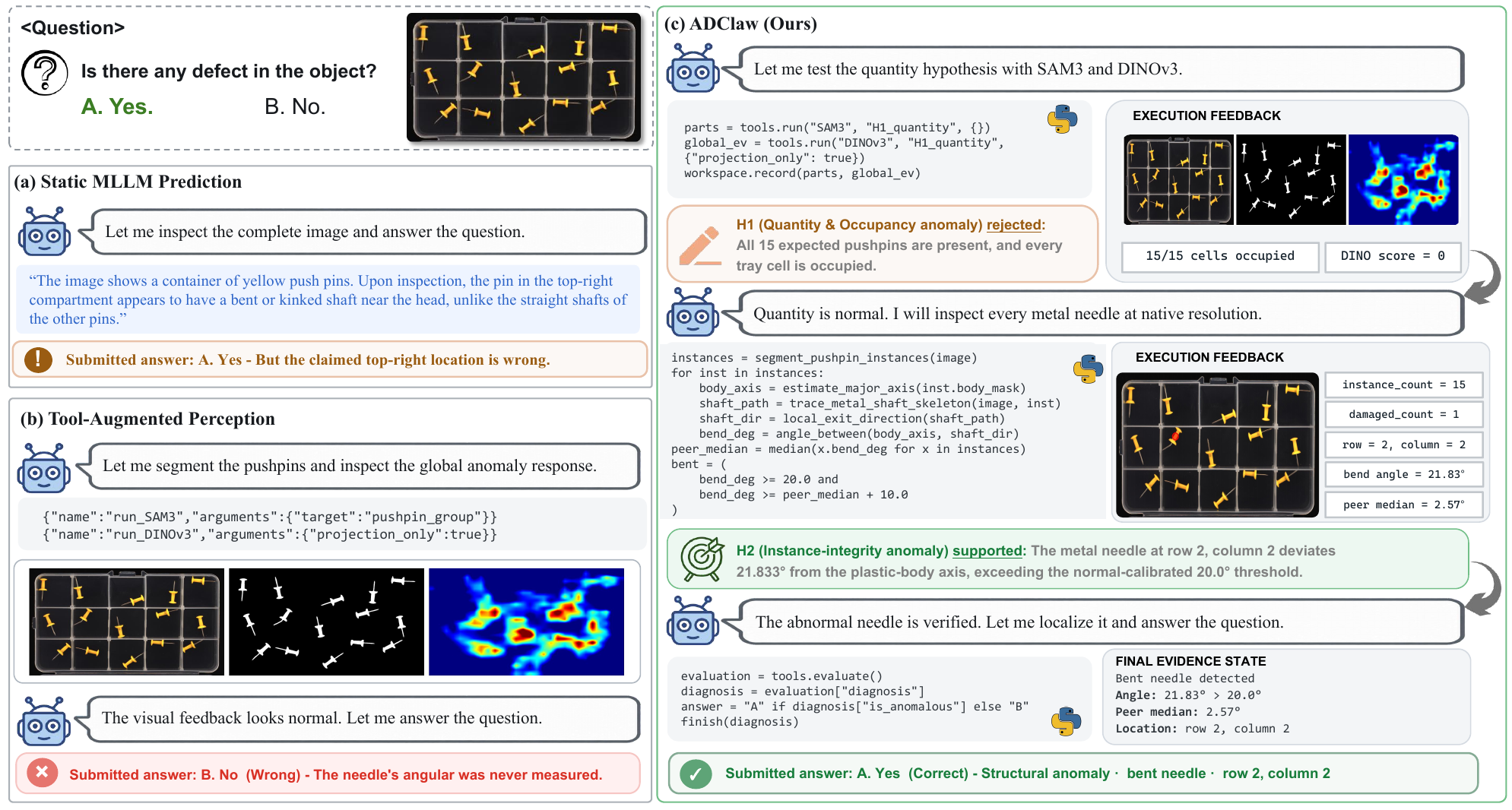}
    \caption{
    \papertext{\textbf{Comparison of inspection strategies on an MVTec LOCO AD example.}
    (a) A single-pass VLM identifies a bent shaft but mislocalizes it.
    (b) Tool-augmented perception incorrectly predicts normality.
    (c) ADClaw tests anomaly hypotheses through executable code,
    checking occupancy and shaft alignment to locate the bent pushpin
    in row 2, column 2.
    }
    {\textbf{MVTec LOCO AD 图钉托盘的检测过程。}
    (a) 单次推理的视觉语言模型识别出针杆弯曲，但定位有误。
    (b) 使用分割结果和全图异常响应图的方案将样本判为正常。
    (c) ADClaw 核对格位占用情况后测量针杆的偏转角，将弯曲图钉定位在第 2 行第 2 列。}
    }
    \label{fig:inspection_comparison}
\end{figure*}

\papertext{Industrial anomaly detection (IAD) aims to identify and localize deviations from normal product conditions and is essential to automated quality inspection in manufacturing~\citep{batzner2024efficientad,liu2023simplenet}. A defect may occupy only a small fraction of an image~\citep{bergmann2020uninformed}, and changes in illumination can make a normal surface appear anomalous~\citep{wang2024realiad,hecklerkram2025mvtecad2}. Detecting a deformation can require a geometric measurement, while identifying a misplaced component requires examining its relationship to the rest of the assembly~\citep{wang2023m3dm,bergmann2022loco,gu2025univad,zhang2025logsad}. Consequently, a region that looks unusual in isolation may be normal for that part, and an intact part can still be in the wrong place. Reliable diagnosis requires identifying which property to examine, where and at what scale to measure it, and what constitutes a normal value or arrangement for the product.}
{工业异常检测（IAD）旨在识别并定位产品中偏离正常状态之处，对制造业的自动化质量检测至关重要~\citep{batzner2024efficientad,liu2023simplenet}。缺陷可能仅占图像中很小的一部分~\citep{bergmann2020uninformed}，光照变化也可能使正常表面看起来异常~\citep{wang2024realiad,hecklerkram2025mvtecad2}。识别形变可能需要测量部件的几何形态，而发现部件错位则需要考察它与装配体其余部分的关系~\citep{wang2023m3dm,bergmann2022loco,gu2025univad,zhang2025logsad}。因此，单独看起来可疑的区域可能属于该部件的正常变化，外观完好的部件也可能被装在错误的位置。可靠的诊断需要确定应检查哪一项属性、在何处、以何种尺度测量，以及该产品正常的测量值或部件关系。}

\papertext{This distinction between locating a deviation and interpreting it is reflected in the evolution of IAD benchmarks. MVTec AD and VisA pair image-level labels with pixel-level defect masks, while MVTec 3D-AD and Real3D-AD extend localization to 3D scans~\citep{bergmann2019mvtec,zou2022visa,bergmann2022mvtec3d,liu2023real3d}. More recent benchmarks ask what the localized evidence means~\citep{fan2025manta}. VisA-D\&R asks models to describe defects and suggest possible causes, and MMAD includes questions on defect type, location, and analysis among the seven tasks it evaluates~\citep{xu2025anomalyov,jiang2025mmad}. Through image--text supervision, vision-language models (VLMs) can connect visual features to defect descriptions~\citep{radford2021clip,li2023blip2} and incorporate observations from specialist detectors to identify anomalies~\citep{gu2024anomalygpt,li2026myriad}. Datasets such as MMR-AD further provide reasoning annotations to support domain-specific training for anomaly understanding~\citep{yao2026mmrad}. Many methods adapt VLMs through supervised fine-tuning, sometimes combined with reinforcement learning~\citep{xu2025anomalyov,chao2025anomalyr1,li2026iadr1}, yet precise localization remains challenging~\citep{zhang2026indomain}. In single-pass inference, a model that encounters an ambiguous region must answer from its initial observation, with no opportunity to inspect it more closely or compare it with the corresponding part in a normal sample of the same product at a comparable scale.}
{从定位偏差到解释偏差，这一变化也体现在 IAD 基准的发展中。MVTec AD 和 VisA 同时提供图像级标签与像素级缺陷掩码，MVTec 3D-AD 和 Real3D-AD 则将定位任务扩展到三维扫描~\citep{bergmann2019mvtec,zou2022visa,bergmann2022mvtec3d,liu2023real3d}。较新的基准进一步考察模型能否解释局部证据的含义~\citep{fan2025manta}。VisA-D\&R 要求模型描述缺陷并分析其可能成因，MMAD 的七项任务则包含缺陷分类、定位与分析~\citep{xu2025anomalyov,jiang2025mmad}。通过图文监督，视觉语言模型（VLM）能够将视觉特征与缺陷描述联系起来~\citep{radford2021clip,li2023blip2}，并借助专用检测器提供的观测识别异常~\citep{gu2024anomalygpt,li2026myriad}。MMR-AD 等数据集进一步提供推理过程标注，为模型学习异常分析提供训练资源~\citep{yao2026mmrad}。许多方法通过监督微调或结合强化学习进行领域适配~\citep{xu2025anomalyov,chao2025anomalyr1,li2026iadr1}，但精确定位仍然具有挑战性~\citep{zhang2026indomain}。在单次推理中，模型遇到存疑区域时必须依据初始观测作答，无法进一步复查该区域，也无法再以相近尺度将其与同种产品正常样本中的对应部件进行比较。}

\papertext{To revisit uncertain predictions, tool-using inspection agents can zoom in on a region~\citep{wu2024vstar}, retrieve normal examples, or consult product knowledge~\citep{miao2026agentiad,zhang2026anomalyagent,zhang2026adseeker}. They can also test candidate defects against normal references over several rounds and revise a judgment as evidence accumulates~\citep{jiang2026anomalyclaw}. With a predefined tool interface, however, the agent can request only the operations that the interface exposes. Testing a hypothesis may require new computations on the outputs of earlier tool calls. If a diagnosis requires combining a part mask with an earlier anomaly map and measuring how far that part deviates from a matched normal reference, selecting the next tool may leave the relevant property unmeasured. An untested hypothesis can lead to a missed defect, as Figure~\ref{fig:inspection_comparison} illustrates, or to an incorrect diagnosis of its type. \emph{How, then, can an inspection agent use what it has already observed to construct the next check needed to test a specific anomaly hypothesis?}}
{为复核存疑的预测，使用工具的检测智能体可以放大局部区域~\citep{wu2024vstar}、检索正常样本或查询产品知识~\citep{miao2026agentiad,zhang2026anomalyagent,zhang2026adseeker}。智能体还可以围绕候选缺陷进行多轮正常参考比对，并随着证据积累修正判断~\citep{jiang2026anomalyclaw}。然而，预定义的工具接口限制了智能体能够直接请求的操作。检验假设可能需要对先前工具调用的输出开展新的计算。若诊断需要将部件掩码与先前得到的异常图结合，测量该部件相对于匹配的正常参考偏离了多少，仅选择下一个工具可能仍无法测得所需属性。假设未得到充分检验，可能导致图~\ref{fig:inspection_comparison} 所示的漏检，也可能导致缺陷类型误判。\emph{那么，检测智能体如何利用已有观测，构造检验当前异常假设所需的下一项检查？}}

\Needspace{2\baselineskip}
\papertext{We approach this problem by letting the agent express each inspection action as executable code. Given a suspected defect, the agent writes a check that combines perception tools with computations on their earlier outputs. The execution result supplies a new observation and determines what remains to be checked. A product-specific normality rule or a matching normal reference gives the measurement its diagnostic meaning. This makes it possible to revise either the measurement or the hypothesis when an observation is inconclusive or conflicts with a measurement from an earlier step. We implement this approach in \textbf{ADClaw}, an industrial inspection agent that keeps its pretrained models fixed and constructs these checks at inference time for the product and region under inspection.}
{我们的思路是让智能体以可执行代码表达每一步检测动作。针对疑似缺陷，智能体编写检查程序，调用感知工具，并对先前的工具输出继续运算。执行结果形成新的观测，据此可确定还有什么需要检查。产品特定的正常性规则或与目标匹配的正常参考，为解释该测量结果提供判据。如果当前观测尚不足以作出诊断，或与先前的测量结果矛盾，智能体便可以调整测量方式或修正异常假设。基于这一思路，我们提出工业检测智能体 \textbf{ADClaw}。它保持预训练模型参数固定，在推理时针对当前产品及待检区域构造检查。}

\papertext{ADClaw begins each check with a \textbf{Structured Normality Prior} that links objects and parts to applicable rules and retrieves normal visual references at the relevant scale. Guided by this prior, the agent writes and executes code in a persistent \textbf{Inspection Workbench}. The workbench retains masks, anomaly maps, and measurements together with their inspection targets, spatial extent, and sources. A later action can restrict a full-image response to a newly identified part in 2D or use a geometric measurement to select a complementary view of the same 3D scan. After execution, \textbf{Hypothesis-Guided Evidence Reasoning} checks that the observations required by the hypothesis are present and assesses agreement among the applicable scores, locations, and defect labels. Missing or conflicting evidence guides the next code action. When inspection ends, the agent fixes its decision and localization before generating an explanation. Our contributions are as follows.}
{每项检查首先由\textbf{结构化正常性先验}将待检对象及部件与适用规则关联起来，并检索尺度匹配的正常视觉参考。在该先验指导下，智能体在持久化的\textbf{检测工作台}中编写并执行代码。工作台保留掩码、异常响应图和测量结果，同时记录各项结果的检测目标、空间范围及来源。由此，在二维检测中，后续动作可从已有的全图响应中提取新识别部件的局部响应。在三维检测中，智能体则可依据几何测量为同一次扫描选择互补视图。执行后，\textbf{假设引导的证据推理}检查检验该假设所需的观测是否齐备，并在适用时核验异常分数、空间位置与缺陷类别的一致性。尚缺或相互冲突的证据指导下一步代码动作。检测结束后，智能体先确定判定与定位结果，再生成解释。本文的主要贡献如下。}
\begin{itemize}
\item \papertext{We cast each inspection step as a code action that tests a suspected defect. The agent uses the result of one check to design the next, reusing earlier measurements to refine its diagnosis without task-specific fine-tuning.}{我们用可执行代码表达每一步检验疑似缺陷的动作。智能体根据本次检查的结果设计下一项检查，并复用已有测量来修正诊断，整个过程无需任务专用微调。}
\item \papertext{The Structured Normality Prior links a part to its rules. The Inspection Workbench runs the check and saves its results. Hypothesis-Guided Evidence Reasoning tests if they support the suspected defect or warrant a new check.}{结构化正常性先验将待检部件与适用规则关联起来，检测工作台执行检查并保存结果，假设引导的证据推理则判断结果是否支持疑似缺陷，或是否需要追加检查。}
\item \papertext{ADClaw achieves competitive results on 2D and 3D defect detection and localization. On MMAD, it also improves on adaptive tool calling in identifying a defect's type and location with the same backbone and tools.}{ADClaw 在二维和三维工业基准的缺陷检测与定位任务中取得了有竞争力的结果。在 MMAD 上，使用相同骨干模型和工具时，ADClaw 对缺陷类型和位置的判断比自适应工具调用基线更准确。}
\end{itemize}

\section{\papertext{Related Work}{相关工作}}

\begin{figure*}[t]
    \centering
    \includegraphics[width=\linewidth]{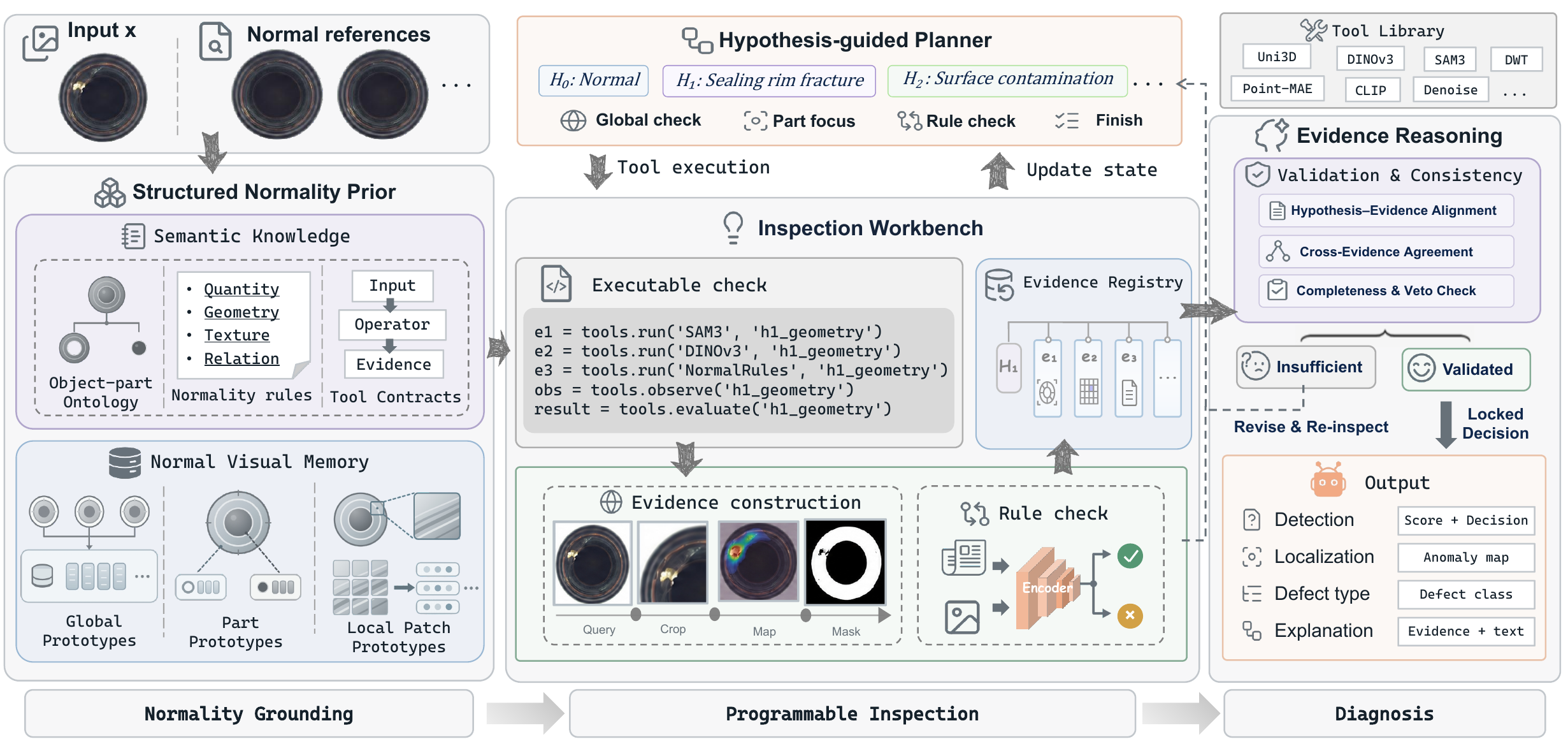}
    \caption{\papertext{\textbf{Overview of the ADClaw framework.}
    The Structured Normality Prior guides inspection.
    The Inspection Workbench executes code and reuses evidence.
    Hypothesis-Guided Evidence Reasoning validates evidence for diagnosis.
    }
    {\textbf{ADClaw 的检测循环。}产品规则与正常参考指导每一步代码动作。检测工作台保存该动作的测量结果及其来源；规划器据此接受或修正假设，或编写代码进一步核查。}}
    \label{fig:framework}
\end{figure*}

\papertext{\textbf{Industrial Anomaly Detection and Understanding.}
Many industrial detectors learn what an intact product looks like from normal samples~\citep{zavrtanik2021draem,gudovskiy2022cflow,defard2021padim}. They localize deviations by comparing a test patch with stored normal features~\citep{roth2022patchcore,damm2025anomalydino} or by measuring disagreement between teacher and student representations~\citep{deng2022reverse,tien2023rdpp,rudolph2023ast}. A misplaced part can nevertheless match a normal patch. To detect such violations, component-level approaches measure part properties and model their relationships~\citep{bergmann2022loco,liu2023componentaware,gu2025univad}. For an assembly, this can mean checking whether all required parts are present and whether each occupies an allowed position~\citep{bergmann2022loco,zhang2025logsad}. In 3D inspection, signals from shape and appearance provide complementary evidence~\citep{wang2023m3dm,cao2024cpmf}. Recent benchmarks also ask models to classify localized defects and explain their visible characteristics~\citep{jiang2025mmad,xu2025anomalyov}.}
{\textbf{工业异常检测与推理。}许多工业检测方法从正常样本中学习完好产品的外观~\citep{zavrtanik2021draem,gudovskiy2022cflow,defard2021padim}，通过比较待检图像块与正常特征库~\citep{roth2022patchcore,damm2025anomalydino}，或度量教师模型与学生模型表征之间的差异，来定位异常~\citep{deng2022reverse,tien2023rdpp,rudolph2023ast}。然而，位置错误的部件在局部仍可能与正常样本相匹配。为识别这类异常，部件级方法测量各部件的属性，并对部件间的关系建模~\citep{bergmann2022loco,liu2023componentaware,gu2025univad}。对于装配体，这意味着需要核查必需的部件是否齐全，以及各部件是否位于允许的位置~\citep{bergmann2022loco,zhang2025logsad}。在三维检测中，形状与外观信息可提供互补证据~\citep{wang2023m3dm,cao2024cpmf}。近期基准还考察模型能否对已定位缺陷分类，并解释其可见特征~\citep{jiang2025mmad,xu2025anomalyov}。}
\vspace{0.2em}

\papertext{\textbf{VLM-Based Industrial Anomaly Detection.}
Vision-language models use text to specify what a detector should look for~\citep{radford2021clip}. Comparing local image features with prompts describing normal and anomalous states enables zero-shot localization~\citep{jeong2023winclip,zhou2024anomalyclip,cao2024adaclip,qu2024vcpclip,ma2025aaclip}. Generative models extend this comparison to defect description through image--text supervision, visual expert features, and instruction tuning~\citep{gu2024anomalygpt,li2026myriad,xu2025anomalyov,liu2023llava}. One way to train a model to describe a local defect is to pair a simulated anomaly with text that specifies what changed and where~\citep{gu2024anomalygpt}. Supervised reasoning traces and reinforcement learning further adapt these models to industrial questions~\citep{li2026iadr1,yao2026mmrad}. Fine-grained localization nevertheless remains challenging after in-domain training~\citep{zhang2026indomain}. A single-pass prediction also gives the model no opportunity to request a new measurement to distinguish a subtle defect from normal variation in the surface appearance of an intact part under varying lighting conditions.}
{\textbf{基于视觉语言模型的工业异常检测。}视觉语言模型通过文本描述指定检测目标~\citep{radford2021clip}。将局部图像特征与描述正常及异常状态的文本提示进行匹配，可以实现零样本定位~\citep{jeong2023winclip,zhou2024anomalyclip,cao2024adaclip,qu2024vcpclip,ma2025aaclip}。生成式模型进一步借助图文监督、视觉专家特征和指令微调来描述缺陷~\citep{gu2024anomalygpt,li2026myriad,xu2025anomalyov,liu2023llava}。为了让模型学会描述局部缺陷，一种训练方式是为模拟异常配上文字，指出发生了何种变化及其所在位置~\citep{gu2024anomalygpt}。推理轨迹监督和强化学习使这些模型进一步适配工业问答任务~\citep{li2026iadr1,yao2026mmrad}。即便经过领域内训练，细粒度定位仍有困难~\citep{zhang2026indomain}。当初始观测存在歧义时，单次预测也无法让模型追加测量，以区分细微缺陷与光照条件变化下完好部件表面外观的正常变化。}

\vspace{0.2em}
\papertext{\textbf{Agentic Industrial Anomaly Detection.}
Inspection agents can revisit uncertain regions by cropping or enhancing an image, comparing parts, and retrieving normal examples~\citep{miao2026agentiad,tan2026indusagent,zhang2026anomalyagent}. Policies trained on tool feedback can weigh the diagnostic value of another call against its cost~\citep{tan2026indusagent}. Other agents explicitly try to refute each suspected anomaly against normal references and update their verdict after each tool call~\citep{jiang2026anomalyclaw}. These agents can choose a new tool as evidence changes~\citep{yao2022react}, but a predefined interface constrains which measurements they can request. Local cropping helps examine suspicious regions, but quantifying their deviation from normality requires measurements within the target region and comparison with matched normal references. ADClaw lets the agent write a follow-up check that operates directly on earlier masks, maps, and measurements. A check measures part-level deviations against matched normal references and preserves each measurement and its source region for later checks.}
{\textbf{基于智能体的异常检测。}检测智能体可通过裁剪或增强图像、比较部件及检索正常样本，复查存疑区域~\citep{miao2026agentiad,tan2026indusagent,zhang2026anomalyagent}。利用工具反馈训练的策略可在追加调用的诊断价值与成本之间作出权衡~\citep{tan2026indusagent}。另一些方法则依据正常参考尝试反驳各个异常假设，在每次工具调用后更新判断~\citep{jiang2026anomalyclaw}。这些智能体可以随证据变化选择新工具~\citep{yao2022react}，但预定义接口限制了它们能够请求的测量。局部裁剪有助于观察可疑区域，但量化其相对正常状态的偏差，还需结合目标范围进行测量，并与匹配的正常参考比较。ADClaw 允许智能体编写后续检查程序，直接对已有掩码、异常图和测量结果继续运算。检查程序将部件级偏差与匹配的正常参考比较，并保留每项测量及其来源区域，供后续检查使用。}
\vspace{-0.5em}
\section{\papertext{Method}{方法}}

\papertext{Given a product image or 3D scan $x$ of category $c$, ADClaw tests anomaly hypotheses through executable code (Figure~\ref{fig:framework}). The \textbf{Structured Normality Prior} (SNP) supplies rules and normal references for the suspected deviation. Guided by this prior, the planner writes actions in the persistent \textbf{Inspection Workbench} (IW), combining tool calls with computations on retained outputs. \textbf{Hypothesis-Guided Evidence Reasoning} (HER) checks whether the results support the defect and its location, and directs further inspection when evidence is incomplete or inconsistent. Models remain frozen, with normal references supporting comparison and calibration.}{给定类别 $c$ 的产品图像或三维扫描 $x$，ADClaw 通过可执行代码检验异常假设（图~\ref{fig:framework}）。\textbf{结构化正常性先验}（SNP）为疑似偏差提供规则和正常参考。规划器据此在持久化\textbf{检测工作台}（IW）中编写动作，将工具调用与已有输出上的计算结合。\textbf{假设引导的证据推理}（HER）检验结果是否支持缺陷及其位置，并在证据不足或矛盾时指导进一步检查。模型保持冻结，正常参考用于比较和校准。}

\subsection{\papertext{Structured Normality Prior}{结构化正常性先验}}
\label{sec:normality_prior}

\papertext{For category $c$, the Structured Normality Prior combines semantic
constraints and normal visual references:}{对于类别 $c$，结构化正常性先验结合语义约束与正常视觉参考：}
\begin{equation}
    \mathcal{P}_c = (\mathcal{K}_c,\mathcal{M}_c),
    \qquad
    \mathcal{K}_c = (\mathcal{O}_c,\mathcal{R}_c,\mathcal{T}),
    \label{eq:structured_normality_prior}
\end{equation}
\papertext{where $\mathcal{O}_c$, $\mathcal{R}_c$, and $\mathcal{T}$ denote
the object--part ontology, normality rules, and tool contracts,
respectively. The visual memory $\mathcal{M}_c$ is built
from the normal reference set $\mathcal{N}_c$.}{其中，$\mathcal{O}_c$、$\mathcal{R}_c$ 和 $\mathcal{T}$ 分别表示对象—部件本体、正常性规则和工具契约。视觉记忆 $\mathcal{M}_c$ 由正常参考集 $\mathcal{N}_c$ 构建。}
\vspace{-1em}
\paragraph{\papertext{Semantic knowledge.}{语义知识。}}
\papertext{SNP organizes normality knowledge through an object--part hierarchy linked to multi-scale visual memory. The ontology $\mathcal O_c$ specifies observable part attributes, inter-part relations, and grounding prompts. Normality rules $\mathcal R_c$ associate the corresponding constraints with inspection targets, required observations, and tool operations. Tool contracts $\mathcal T$ define inputs, outputs, and applicability. Semantic information is manually organized and reviewed to provide explicit inspection targets and criteria.}{SNP 通过对象—部件层级与多尺度视觉记忆组织正常性知识。本体 $\mathcal O_c$ 规定部件的可观测属性、相互关系与视觉定位提示。正常性规则 $\mathcal R_c$ 将相关约束关联到检查目标、所需观测和工具操作。工具契约 $\mathcal T$ 定义输入、输出及适用条件。上述语义信息经过人工整理与确认，为后续检查提供明确的目标及判据。}
\vspace{-1em}
\paragraph{\papertext{Normal visual memory.}{正常视觉记忆。}}
\papertext{Frozen encoders construct a \emph{semantic-topological hierarchical memory} from normal references at global, part, and local scales. Its organization is defined in Equation~\eqref{eq:hierarchical_normal_memory}:}{冻结编码器从正常参考中构建包含全局、部件和局部三个尺度的\emph{语义—拓扑分层记忆}，其组织形式由式~\eqref{eq:hierarchical_normal_memory} 定义：}
\begin{equation}
    \mathcal M_c = \bigl(
    \mathcal M_c^{\mathrm g},
    \{\mathcal M_{c,p}^{\mathrm p}\}_{p\in\Pi_c},
    \mathcal M_c^{\mathrm l},
    \mathcal A_c
    \bigr).
    \label{eq:hierarchical_normal_memory}
\end{equation}
\papertext{Here, $\mathcal M_c^{\mathrm g}$ stores global normal features and prototypes that represent overall appearance modes. $\Pi_c$ indexes semantic parts with available reference memory, and $\mathcal M_{c,p}^{\mathrm p}$ stores the normal features and prototypes of part $p$. Local features and prototypes in $\mathcal M_c^{\mathrm l}$ support fine-grained deviation comparison. $\mathcal A_c$ stores cluster layouts, adjacency statistics, and available geometric relations between parts, capturing the spatial organization of normal structures. Shared category, part, and sample identifiers link these references and relations, providing normality references at the relevant scale. Semantic configurations and visual memory are established before inspection and reused across samples.}{其中，$\mathcal M_c^{\mathrm g}$ 保存全局正常特征及原型，表征整体外观模式。$\Pi_c$ 表示已建立参考记忆的语义部件集合，$\mathcal M_{c,p}^{\mathrm p}$ 保存部件 $p$ 的正常特征及原型。$\mathcal M_c^{\mathrm l}$ 中的局部特征与原型支持细粒度偏差比较。$\mathcal A_c$ 保存聚类布局、邻接统计及可用的部件几何关系信息，刻画正常结构的空间组织。共享的类别、部件与样本标识将这些参考及其关系关联起来，使检查能够调用相应尺度的正常性依据。语义配置与视觉记忆在检查前建立，并在后续样本检查中复用。}

\papertext{For 2D nearest-neighbor comparison, let
$\hat{\mathbf z}=\mathbf z/\|\mathbf z\|_2$ denote a normalized
query descriptor and $\mathcal{B}$ the corresponding set of
normalized normal features. The feature deviation is}{对于二维最近邻比较，令 $\hat{\mathbf z}=\mathbf z/\|\mathbf z\|_2$ 表示归一化查询描述子，$\mathcal{B}$ 表示对应的归一化正常特征集合。特征偏差为}
\begin{equation}
    d(\mathbf z,\mathcal{B})
    =
    \operatorname{clip}_{[0,1]}
    \left(
        1-\max_{\mathbf b\in\mathcal{B}}
        \hat{\mathbf z}^{\top}\mathbf b
    \right),
    \label{eq:normal_feature_distance}
\end{equation}
\papertext{where $\operatorname{clip}_{[0,1]}$ clips values to $[0,1]$.
Global and part deviations use their corresponding prototypes.
Target-conditioned patch comparisons use retrieved normal features,
whereas dense responses use the full category-specific normal patch memory.}{其中，$\operatorname{clip}_{[0,1]}$ 将数值截断到 $[0,1]$。全局与部件偏差使用各自对应的原型。以目标为条件的图像块比较使用检索到的正常特征，稠密响应则使用完整的类别专用正常图像块记忆。}

\papertext{For 3D inspection, memory stores normal point-cloud templates for registered geometric comparison and frozen features for feature matching. Normal-sample statistics calibrate each scoring branch.}{三维记忆保存用于配准后几何比较的正常点云模板，以及用于特征匹配的冻结特征。正常样本统计量校准各评分分支。}

\subsection{\papertext{Inspection Workbench}{检测工作台}}
\label{sec:active_inspection}

\papertext{Given input $x$, prior $\mathcal P_c$, and persistent evidence state $\mathcal E_t$, the planner writes action $a_t$ to address the current hypothesis's evidence requirements. Execution updates the state as}{给定输入 $x$、先验 $\mathcal P_c$ 和持久化证据状态 $\mathcal E_t$，规划器编写动作 $a_t$ 以满足当前假设的证据需求。执行后状态更新为}
\begin{equation}
    \mathcal E_{t+1}
    =
    \mathcal E_t \cup
    \operatorname{Execute}(a_t;x,\mathcal P_c,\mathcal E_t).
    \label{eq:evidence_state_update}
\end{equation}

\papertext{Tool outputs are registered in the Evidence Registry, with their
key contents summarized as}{工具输出登记到证据注册表中，其关键内容概括为}
\begin{equation}
    e =
    \bigl(
    \mathrm{id}_e,h_e,r_e,\tau_e,
    s_e^{\mathrm{raw}},\widetilde s_e,
    q_e,\Omega_e,\mathrm{provenance}_e
    \bigr).
    \label{eq:evidence_record}
\end{equation}
\papertext{Each record links its identifier $\mathrm{id}_e$ to hypothesis $h_e$, applicable rules $r_e$, and tool $\tau_e$. It stores the raw score $s_e^{\mathrm{raw}}$, calibrated score $\widetilde s_e$, quality $q_e$, spatial scope $\Omega_e$, and source and upstream IDs in $\mathrm{provenance}_e$. Rule associations reside in hypothesis or tool metadata. Numerical fields may be undefined for spatial or semantic outputs. Masks, responses, and references remain reusable, with dependencies linking derived measurements to their inputs.}{每条记录将标识 $\mathrm{id}_e$ 与假设 $h_e$、适用规则 $r_e$ 及工具 $\tau_e$ 关联，并保存原始分数 $s_e^{\mathrm{raw}}$、校准分数 $\widetilde s_e$、质量 $q_e$、空间范围 $\Omega_e$，以及 $\mathrm{provenance}_e$ 中的来源和上游标识。规则关联存于假设或工具元数据。空间或语义输出的数值字段可以未定义。掩码、响应与参考可供复用，依赖关系将派生测量与其输入关联。}

\papertext{In 2D, restricting the shared full-image response $A$ to target $\Omega_h$ yields $A|_{\Omega_h}$ for composition, structural, and semantic checks. In 3D, geometric evidence guides complementary virtual views of anomalous regions. HER assesses results and returns unresolved requirements to the planner.}{二维检测将共享全图响应 $A$ 限制至目标 $\Omega_h$，得到用于组成、结构和语义检查的 $A|_{\Omega_h}$。三维检测依据几何证据选择异常区域的互补虚拟视图。HER 评估结果，并向规划器返回未解决的证据需求。}

\begin{table*}[t]
\centering
\setlength{\abovecaptionskip}{0pt}
\caption{
\papertext{\textbf{2D anomaly detection performance on
MVTec AD, VisA, MVTec LOCO AD, and GoodsAD}.
I-R and P-R denote image- and pixel-level AUROC; F1 denotes F1-max.
All scores are percentages.
\textbf{Bold} and \underline{underlining} mark the best and second-best results.}{\textbf{MVTec AD、VisA、MVTec LOCO AD 和 GoodsAD 上的二维异常检测性能}。I-R 和 P-R 分别表示图像级与像素级 AUROC；F1 表示 F1-max。所有分数均以百分比报告。\textbf{粗体}和\underline{下划线}分别标记最优与次优结果。}
}
\vspace{0.2em}
\label{tab:2d_compare}

\setlength{\tabcolsep}{\papertext{8.38pt}{6pt}}
\renewcommand{\arraystretch}{1.08}

\footnotesize
\begin{tabular}{lcccccccccccc}
\toprule
\multirow{2}{*}{\papertext{Method}{方法}}
& \multicolumn{3}{c}{MVTec AD}
& \multicolumn{3}{c}{VisA}
& \multicolumn{3}{c}{MVTec LOCO AD}
& \multicolumn{3}{c}{GoodsAD} \\

\cmidrule(lr){2-4}
\cmidrule(lr){5-7}
\cmidrule(lr){8-10}
\cmidrule(lr){11-13}

&
I-R$\uparrow$ & P-R$\uparrow$ & F1$\uparrow$ &
I-R$\uparrow$ & P-R$\uparrow$ & F1$\uparrow$ &
I-R$\uparrow$ & P-R$\uparrow$ & F1$\uparrow$ &
I-R$\uparrow$ & P-R$\uparrow$ & F1$\uparrow$ \\

\midrule

\rowcolor{black!6}
\multicolumn{13}{l}{\textbf{\papertext{\textit{Traditional IAD}}{传统工业异常检测}}} \\
PatchCore~\citep{roth2022patchcore}
& \textbf{98.46} & \underline{98.07} & \underline{98.04}
& \underline{90.52} & \underline{97.99} & \underline{88.36}
& 80.02 & \textbf{87.97} & 81.47
& 72.03 & \underline{96.66} & 75.85 \\

FastFlow~\citep{yu2021fastflow}
& 94.01 & 97.69 & 93.91
& 87.45 & 97.09 & 86.95
& \underline{81.34} & 82.83 & \underline{82.40}
& \underline{72.97} & 89.17 & \underline{76.42} \\

\midrule

\rowcolor{black!6}
\multicolumn{13}{l}{\textbf{\papertext{\textit{VLM-based IAD}}{基于 VLM 的工业异常检测}}} \\
WinCLIP~\citep{jeong2023winclip}
& 70.11 & 80.97 & 86.14
& 64.26 & 70.05 & 74.23
& 53.47 & 52.87 & 77.36
& 54.59 & 79.81 & 71.24 \\

AnomalyCLIP~\citep{zhou2024anomalyclip}
& 91.55 & 91.07 & 92.75
& 81.97 & 95.47 & 80.44
& 61.86 & 63.75 & 77.33
& 57.25 & 83.49 & 71.39 \\

AnomalyR1~\citep{chao2025anomalyr1}
& 83.66 & -- & 89.50
& 77.37 & -- & 78.64
& 58.32 & -- & 77.49
& 58.32 & -- & 71.50 \\

Qwen3.5-9B~\citep{qwen35}
& 95.69 & -- & 95.41
& 87.92 & -- & 84.28
& 71.11 & -- & 79.23
& 53.53 & -- & 71.80 \\

\cmidrule(lr){1-13}

\rowcolor{black!6}
\textbf{ADClaw}
& \underline{97.68} & \textbf{98.22} & \textbf{98.33}
& \textbf{95.26} & \textbf{98.99} & \textbf{92.57}
& \textbf{84.10} & \underline{85.73} & \textbf{83.22}
& \textbf{82.94} & \textbf{97.01} & \textbf{81.35} \\

\bottomrule
\end{tabular}
\end{table*}

\subsection{\papertext{Hypothesis-Guided Evidence Reasoning}{假设引导的证据推理}}
\label{sec:evidence_reasoning}

\papertext{HER forms $\mathcal E_h$ from the latest eligible scored record of each applicable tool for hypothesis $h$. Eligible records require successful execution, a valid calibrated score, and positive quality. Decision evidence $\mathcal D_h\subseteq\mathcal E_h$ contributes to the hypothesis score}{HER 使用各适用工具最新的合格评分记录构成假设 $h$ 的证据集合 $\mathcal E_h$。合格记录要求执行成功、具有有效校准分数且质量为正。用于数值评分的决策证据构成 $\mathcal D_h\subseteq\mathcal E_h$，假设分数为}
\begin{equation}
s_h =
\frac{
    \sum_{e \in \mathcal{D}_h} w_e q_e \widetilde{s}_e
}{
    \sum_{e \in \mathcal{D}_h} w_e q_e
}.
\label{eq:hypothesis_evidence_score}
\end{equation}
\papertext{Here, $\widetilde s_e$ is the anomaly score calibrated against normal references, $q_e\in[0,1]$ is the validity or confidence indicator returned by the tool, and $w_e>0$ is its fixed tool-specific weight. These weights apply only to $\mathcal D_h$; with one decision record, $s_h=\widetilde s_e$. An empty decision set yields insufficient evidence. Auxiliary observations provide rule, spatial, or semantic support and conditional vetoes; unscored artifacts remain available for subsequent checks.}{其中，$\widetilde s_e$ 为依据正常参考校准的异常分数，$q_e\in[0,1]$ 为工具返回的有效性或置信指标，$w_e>0$ 为该工具的固定权重。权重仅作用于 $\mathcal D_h$；仅有一条决策记录时，$s_h=\widetilde s_e$。决策集合为空时返回证据不足。辅助观测提供规则、空间或语义支持及条件否决，未评分产物仍可用于后续检查。}

\papertext{ADClaw further evaluates \emph{cross-evidence consistency} in score, spatial, and
semantic dimensions:}{ADClaw 进一步在分数、空间和语义三个维度评估\emph{跨证据一致性}：}
\begin{equation}
C_h =
\mathbf{1}\!\left[\sigma_h \leq \tau_{\mathrm{score}}\right]
\mathbf{1}\!\left[\iota_h \geq \tau_{\mathrm{loc}}\right]
\mathbf{1}\!\left[\rho_h \geq \tau_{\mathrm{sem}}\right].
\label{eq:evidence_consistency}
\end{equation}
\papertext{Here, $\sigma_h$ is the standard deviation of decision scores; $\iota_h$ is the minimum IoU between a reference box and the other available boxes; and $\rho_h$ is the dominant-label fraction among available semantic labels. Spatial and semantic factors are evaluated with at least two corresponding observations and otherwise set to one; required observations remain subject to completeness checks. An anomaly hypothesis is supported when evidence, score, consistency, spatial, support, and veto conditions are satisfied}{其中，$\sigma_h$ 为决策分数的标准差；$\iota_h$ 为参考定位框与其余可用定位框之间的最小交并比；$\rho_h$ 为可用语义标签中主导标签的占比。空间与语义因子在至少存在两项对应观测时计算，否则取 $1$；必要观测另由完整性条件检查。支持异常假设要求证据、分数、一致性、空间、辅助支持及否决条件均满足}
\begin{equation}
A_h =
\mathbf{1}\!\left[|\mathcal{E}_h| \geq m\right]
\mathbf{1}\!\left[s_h > \tau_h\right]
C_h G_h R_h (1-V_h).
\label{eq:hypothesis_acceptance}
\end{equation}
\papertext{Here, $m$ is the minimum evidence count and $\tau_h$ the hypothesis threshold. $G_h$, $R_h$, and $V_h$ denote the spatial check, support requirement, and veto condition. Missing or conflicting evidence guides further inspection or hypothesis revision within the budget. ADClaw fixes its score, decision, and localization before generating an explanation that identifies the supporting measurements, their sources, and the applicable normality rules or reference comparisons used to justify the diagnosis. These links preserve the measurements and criteria supporting the final decision.}{其中，$m$ 为最低证据数量，$\tau_h$ 为假设阈值。$G_h$、$R_h$ 和 $V_h$ 分别表示空间检查、支持要求及否决条件。缺失或冲突证据指导预算内的进一步检查或假设修正。ADClaw 先固定分数、决策及定位结果，再生成解释，指明支持判断的测量及其来源，以及支持诊断的正常性规则或参考比较。这些关联保留了最终决策所依据的测量与判据。}

\setlength{\textfloatsep}{10pt plus 3pt minus 2pt}
\setlength{\dbltextfloatsep}{10pt plus 3pt minus 2pt}
\section{\papertext{Experiments}{实验}}

\paragraph{\papertext{Experimental setup.}{实验设置。}}
\papertext{We evaluate 2D anomaly detection on MVTec AD~\citep{bergmann2019mvtec},
VisA~\citep{zou2022visa}, MVTec LOCO AD~\citep{bergmann2022loco},
and GoodsAD~\citep{zhang2024goodsad}, 3D detection on
Real3D-AD~\citep{liu2023real3d} and
MVTec 3D-AD~\citep{bergmann2022mvtec3d}, and anomaly understanding
on MMAD~\citep{jiang2025mmad}.
We report image-level AUROC and F1-max, and pixel-level AUROC
for 2D detection; object-level AUROC and average precision,
and point-level AUROC for 3D detection.
MVTec 3D-AD also reports P-PRO, the normalized area under
the PRO--FPR curve up to 0.30 FPR.
Detection and localization metrics are macro-averaged over categories.
MMAD reports task-level and mean accuracy, using balanced accuracy for anomaly discrimination.}{我们在 MVTec AD、VisA、MVTec LOCO AD 和 GoodsAD 上评估二维异常检测，在 Real3D-AD 和 MVTec 3D-AD 上评估三维异常检测，并在 MMAD（38 个类别）上评估异常理解能力。二维检测报告图像级 AUROC、F1-max 与像素级 AUROC；三维检测报告对象级 AUROC、平均精度与点级 AUROC。MVTec 3D-AD 还报告 P-PRO，即 FPR 不超过 0.30 时 PRO--FPR 曲线下的归一化面积。检测与定位指标均在类别间进行宏平均。MMAD 评估报告七项任务准确率及其平均值，其中异常判别采用平衡准确率。}

\begin{table}[t]
\centering
\footnotesize
\setlength{\abovecaptionskip}{0pt}
\caption{
        \papertext{\textbf{Comparison of 3D industrial anomaly detection performance on
Real3D-AD and MVTec 3D-AD.}
O-R and O-AP denote object-level AUROC and average precision;
P-R denotes point-level AUROC. P-PRO is the normalized area under the
PRO--FPR curve up to 0.30 FPR, reported only on MVTec 3D-AD.
\textbf{Bold} and \underline{underline} mark the best and second-best model scores.}{\textbf{Real3D-AD 与 MVTec 3D-AD 上的三维工业异常检测性能比较。}O-R 和 O-AP 分别表示对象级 AUROC 与平均精度；P-R 表示点级 AUROC。P-PRO 为 FPR 不超过 0.30 时 PRO--FPR 曲线下的归一化面积，仅在 MVTec 3D-AD 上报告。\textbf{粗体}和\underline{下划线}分别表示最优与次优模型结果。}
    }
\label{tab:3d_compare}

\setlength{\tabcolsep}{2pt}
\renewcommand{\arraystretch}{1.10}

\resizebox{\columnwidth}{!}{
\begin{tabular}{lccccccc@{}}
\toprule
\multirow{2}{*}{\papertext{Method}{方法}}
& \multicolumn{3}{c}{Real3D-AD}
& \multicolumn{4}{c}{MVTec 3D-AD} \\
\cmidrule(lr){2-4}
\cmidrule(lr){5-8}
& O-R$\uparrow$
& O-AP$\uparrow$
& P-R$\uparrow$
& O-R$\uparrow$
& O-AP$\uparrow$
& P-R$\uparrow$
& P-PRO$\uparrow$ \\
\midrule

\rowcolor{black!6}
\multicolumn{8}{l}{\textbf{\papertext{\textit{Traditional 3D IAD}}{传统三维工业异常检测}}} \\
BTF~\citep{horwitz2023btf}
& 60.10 & 61.00 & 57.11
& 60.55 & 85.45 & 77.41 & 49.25 \\

PatchCore-FPFH~\citep{roth2022patchcore}
& 58.88 & 56.97 & 62.48
& 80.29 & 93.62 & \textbf{96.80} & 89.02 \\

PatchCore-PointMAE~\citep{roth2022patchcore}
& 58.53 & 60.41 & 64.81
& 70.03 & 87.97 & 75.26 & 65.49 \\

M3DM-PointBERT~\citep{wang2023m3dm}
& 53.76 & 58.38 & 61.82
& 69.99 & 89.34 & 95.07 & 89.43 \\

M3DM-PointMAE~\citep{wang2023m3dm}
& 53.82 & 55.41 & 62.28
& 68.25 & 88.41 & 95.28 & \textbf{90.12} \\

\midrule

\rowcolor{black!6}
\multicolumn{8}{l}{\textbf{\papertext{\textit{Language-guided 3D IAD}}{语言引导的三维工业异常检测}}} \\
BTP~\citep{li2026btp}
& 61.40 & 65.10 & \textbf{84.50}
& -- & -- & -- & -- \\

PointAD~\citep{zhou2024pointad}
& 74.80 & 76.90 & 73.50
& 82.00 & 94.20 & 95.50 & 84.40 \\

GS-CLIP~\citep{deng2026gsclip}
& \underline{76.40} & \underline{77.70} & 76.30
& \underline{83.60} & \underline{96.50} & 96.30 & 86.40 \\

\midrule

\rowcolor{black!6}
\textbf{ADClaw}
& \textbf{79.92}
& \textbf{81.16}
& \underline{83.59}
& \textbf{84.27}
& \textbf{97.11}
& \underline{96.48}
& \underline{89.57} \\

\bottomrule
\end{tabular}
}
\end{table}

\vspace{-0.5em}
\paragraph{\papertext{Baselines and implementation.}{基线与实现。}}
\papertext{We compare ADClaw with traditional and VLM-based IAD methods,
3D detectors, and anomaly-understanding models.
ADClaw uses Qwen3.5-9B~\citep{qwen35} with all pretrained
parameters frozen; normal references support visual memory
and calibration.
The 2D toolkit combines DINOv3 ViT-L/16~\citep{simeoni2025dinov3},
SAM3~\citep{carion2025sam3}, CLIP~\citep{radford2021clip},
and the discrete wavelet transform (DWT).
For 3D inspection, we combine point-cloud registration residuals
with frozen PointMAE~\citep{pang2022pointmae} feature comparisons
as complementary geometric and feature evidence,
using at most six $256\times256$ virtual views per scan.
}{我们将 ADClaw 与传统及基于 VLM 的工业异常检测方法、三维检测器和异常理解模型进行比较。ADClaw 采用 Qwen3.5-9B，所有预训练参数均保持冻结；正常参考用于支持视觉记忆与校准。二维工具集包含 DINOv3 ViT-L/16、SAM3、CLIP 和离散小波变换（DWT）。对于三维检查，我们结合点云配准后的几何残差与冻结的 PointMAE 特征比较，获取互补的几何与特征证据，每次扫描最多使用六个 $256\times256$ 虚拟视图。}

\begin{table*}[t]
\centering
\setlength{\abovecaptionskip}{0pt}
\caption{
        \papertext{\textbf{Comparison of anomaly understanding performance on MMAD (accuracy, \%).}
The seven tasks assess anomaly discrimination, defect understanding, and object understanding across categories. \textbf{Bold} and \underline{underlining} mark the best and second-best scores.}{\textbf{MMAD 上的异常理解性能比较（准确率，\%）。}七项任务评估不同工业类别中的异常判别、缺陷理解和对象理解。\textbf{粗体}和\underline{下划线}分别表示最优与次优结果。}
    }
\label{tab:mmad}
\vspace{0.2em}
\setlength{\tabcolsep}{7.24pt}

\footnotesize
\begin{tabular}{lc|c|cccc|cc|c}
\toprule
\multirow{2}{*}{\papertext{Method}{方法}} &
\multirow{2}{*}{\papertext{Scale}{模型规模}} &
\multicolumn{1}{c|}{\papertext{Anomaly}{异常}} &
\multicolumn{4}{c|}{\papertext{Defect}{缺陷}} &
\multicolumn{2}{c|}{\papertext{Object}{对象}} &
\multirow{2}{*}{\papertext{Average}{平均}} \\
\cmidrule(lr){3-3}
\cmidrule(lr){4-7}
\cmidrule(lr){8-9}
& &
{\fontsize{7}{8}\selectfont \papertext{Discrimination}{判别}} &
{\fontsize{7}{8}\selectfont \papertext{Classification}{分类}} &
{\fontsize{7}{8}\selectfont \papertext{Localization}{定位}} &
{\fontsize{7}{8}\selectfont \papertext{Description}{描述}} &
{\fontsize{7}{8}\selectfont \papertext{Analysis}{分析}} &
{\fontsize{7}{8}\selectfont \papertext{Classification}{分类}} &
{\fontsize{7}{8}\selectfont \papertext{Analysis}{分析}} & \\
\midrule

GPT-4o~\citep{gpt4o} & -- &
68.63 & 65.80 & 55.62 & 73.21 & 83.41 & 94.98 & 82.80 & 74.92 \\
GPT-5-mini~\citep{zhang2026indomain} & -- &
64.10 & 67.40 & 69.10 & 79.00 & \underline{86.70} & 94.00 & 83.40 & 77.70 \\
Gemini-1.5-Pro~\citep{gemini15} & -- &
68.63 & 60.12 & 58.56 & 70.38 & 82.46 & 89.20 & 82.25 & 73.09 \\
Claude-3.5-Sonnet~\citep{claude35sonnet} & -- &
60.14 & 58.34 & 48.81 & 67.13 & 79.11 & 85.19 & 79.83 & 68.36 \\
Qwen3.5-9B~\citep{qwen35} & 9B &
65.03 & 62.64 & 63.53 & 76.82 & 84.79 & 92.87 & \underline{86.93} & 76.09 \\
LLaVA-NeXT~\citep{llavanext} & 34B &
57.92 & 48.79 & 52.87 & 71.34 & 80.28 & 81.12 & 77.80 & 67.16 \\
InternVL2~\citep{internvl2} & 76B &
68.25 & 54.22 & 56.66 & 66.30 & 80.47 & 86.40 & 82.92 & 70.75 \\

\midrule

ADSeeker~\citep{zhang2026adseeker} & 7B &
68.62 & 53.82 & 55.57 & 60.51 & 77.40 & 89.91 & 83.48 & 69.90 \\
IAD-R1~\citep{li2026iadr1} & 7B &
66.62 & 54.15 & 60.95 & 66.53 & 78.55 & 86.90 & 76.11 & 69.97 \\
AnomalyR1~\citep{chao2025anomalyr1} & 3B &
60.62 & 63.56 & 70.14 & \underline{80.47} & 85.28 & 92.48 & 86.15 & 76.96 \\
AnomalyClaw~\citep{jiang2026anomalyclaw} & 9B &
\underline{71.68} & 64.53 & 64.06 & 78.12 &
84.62 & 93.17 & 86.27 & 77.49 \\
OmniAD~\citep{zhao2026omniad} & 7B &
68.80 & \textbf{78.80} & \underline{75.50} & 67.20 &
86.40 & \textbf{96.00} & 86.40 & \underline{79.90} \\

\midrule
\rowcolor{black!6}
\textbf{ADClaw} & 9B &
\textbf{74.98} &
\underline{73.22} &
\textbf{76.15} &
\textbf{80.67} &
\textbf{86.79} &
\underline{95.27} &
\textbf{87.68} &
\textbf{82.11} \\

\bottomrule
\end{tabular}
\end{table*}

\begin{table}[t]
\centering
\captionsetup{position=top,skip=6pt}
\caption{\papertext{\textbf{Key mechanisms on MMAD.} Group mean accuracies (\%) are reported for each variant.}{\textbf{MMAD 上的关键机制。}报告各变体的任务组平均准确率（\%）。}}
\label{tab:core_ablation}
\footnotesize
\setlength{\tabcolsep}{3pt}
\renewcommand{\arraystretch}{1.12}
\fontsize{9}{10.5}\selectfont
\begin{tabularx}{\linewidth}{Xcccc@{}}
\toprule
\papertext{Variant}{变体} & \papertext{Anomaly}{异常} & \papertext{Defect}{缺陷} & \papertext{Object}{对象} & \papertext{Avg.}{平均} \\
\midrule
Qwen3.5-9B & 65.03 & 71.95 & 89.90 & 76.09 \\
\papertext{w/o SNP}{移除 SNP} & 71.80 & 75.85 & 90.70 & 79.51 \\
\papertext{w/o IW}{移除 IW} & 72.90 & 76.65 & 90.90 & 80.19 \\
\papertext{w/o HER}{移除 HER} & 70.80 & 76.70 & 90.90 & 79.91 \\
\midrule
\rowcolor{black!6}
\textbf{\papertext{ADClaw}{ADClaw}} & \textbf{74.98} & \textbf{79.21} & \textbf{91.48} & \textbf{82.11} \\
\bottomrule
\end{tabularx}
\end{table}

\begin{table}[t]
\centering
\captionsetup{position=top,skip=6pt}
\caption{\papertext{\textbf{Inspection strategies on MMAD.} Group means (\%) are compared using the same backbone, tools, and normal references.}{\textbf{MMAD 上的检查策略。}各组使用相同主干、工具和正常参考，报告任务组平均准确率（\%）。}}
\label{tab:inspection_strategy}
\footnotesize
\setlength{\tabcolsep}{3pt}
\renewcommand{\arraystretch}{1.12}
\fontsize{9}{10.5}\selectfont
\begin{tabularx}{\linewidth}{Xcccc@{}}
\toprule
\papertext{Variant}{变体} & \papertext{Anomaly}{异常} & \papertext{Defect}{缺陷} & \papertext{Object}{对象} & \papertext{Avg.}{平均} \\
\midrule
\papertext{Fixed Tool Pipeline}{固定工具流程} & 69.80 & 75.35 & 89.95 & 78.73 \\
\papertext{Adaptive Tool Calling}{自适应工具调用} & 72.40 & 77.15 & 90.63 & 80.32 \\
\midrule
\rowcolor{black!6}
\textbf{\papertext{ADClaw}{ADClaw}} & \textbf{74.98} & \textbf{79.21} & \textbf{91.48} & \textbf{82.11} \\
\bottomrule
\end{tabularx}
\end{table}

\begin{figure*}[t]
    \centering
    \includegraphics[width=\textwidth]{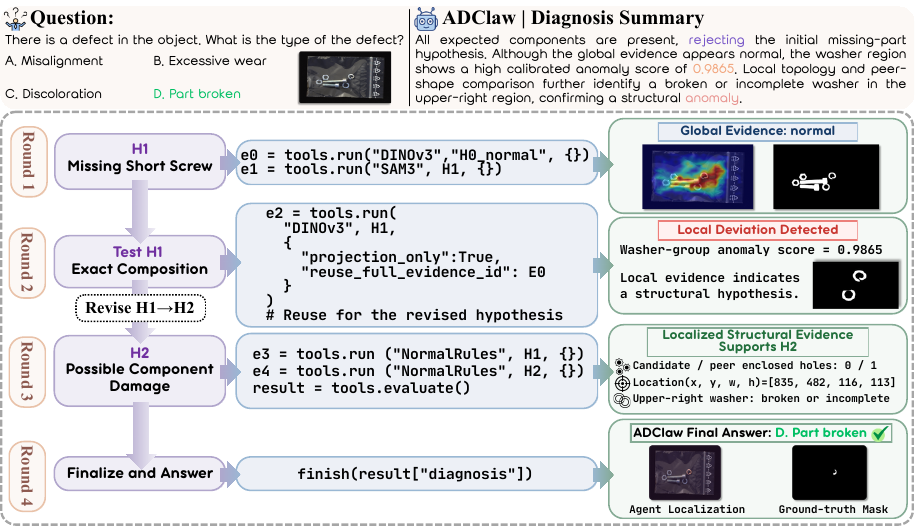}
    \caption{
        \papertext{\textbf{A summarized inspection trace for an MMAD defect-classification example.}
        ADClaw rejects an initial missing-screw hypothesis and identifies a broken
        washer through part-level visual evidence, composition checks, and
        component-integrity evaluation.}{\textbf{MMAD 缺陷分类示例的检查轨迹摘要。}ADClaw 通过部件级视觉证据、组成检查与组件完整性评估，否定初始的螺钉缺失假设，并识别出垫圈破损。}
    }
    \label{fig:mmad_case}
\end{figure*}

\subsection{\papertext{2D Industrial Anomaly Detection}{二维工业异常检测}}
\label{sec:exp_2d}

\papertext{Table~\ref{tab:2d_compare} shows consistent image-level benefits across
2D inspection scenarios: ADClaw leads on seven of eight
image-level metrics, with its largest I-AUROC gains over the Qwen baseline
occurring on MVTec LOCO AD and GoodsAD. This pattern is consistent with
the value of normality-grounded inspection for compositional constraints
and product-specific appearance variations. Normal references, object--part
semantics, and explicit rules jointly contextualize the visual discrepancies
observed during inspection. On MVTec LOCO AD, ADClaw combines the highest image-level AUROC and F1-max with the second-best pixel-level AUROC, supporting effective anomaly discrimination alongside spatially informative inspection. These results support combining visual comparison with structured inspection using frozen pretrained models.}{表~\ref{tab:2d_compare} 展示了 ADClaw 在不同二维检查场景中的一致图像级收益：其在八项图像级指标中的七项取得最优结果，相较于 Qwen 基线的最大 I-AUROC 提升出现在 MVTec LOCO AD 和 GoodsAD 上。这一趋势与基于正常性依据的检查在处理组成约束及产品特定外观变化时的价值相一致。正常参考、对象--部件语义与显式规则共同为检查中观察到的视觉差异提供上下文。在 MVTec LOCO AD 上，ADClaw 同时取得最优图像级 AUROC 与 F1-max，以及次优像素级 AUROC，体现了异常判别与空间定位信息的有效结合。这些结果支持在冻结预训练模型的条件下结合视觉比较与结构化检查。}

\subsection{\papertext{3D Industrial Anomaly Detection}{三维工业异常检测}}
\label{sec:exp_3d}

\papertext{Table~\ref{tab:3d_compare} reports ADClaw's 3D detection and localization on Real3D-AD and MVTec 3D-AD. ADClaw achieves the highest object-level AUROC and average precision on both benchmarks and ranks second on all reported localization metrics, indicating a strong balance between anomaly discrimination and local defect localization. Registration residuals and normal-feature comparisons provide complementary numerical evidence, while programmable inspection links these measurements to specific inspection targets for further hypothesis-specific verification.}{表~\ref{tab:3d_compare} 展示了 ADClaw 在 Real3D-AD 和 MVTec 3D-AD 上的三维检测与定位表现。ADClaw 在两个基准上均取得最高的对象级 AUROC 与平均精度，并在所报告的局部定位指标中均取得次优结果，体现了整体异常判别与局部缺陷定位之间的良好兼顾。配准残差与正常特征比较提供互补的数值证据，可编程检查进一步将这些测量与具体检查目标关联，支持针对异常假设的后续验证。}

\subsection{\papertext{Anomaly Understanding}{异常理解}}
\label{sec:exp_mmad}

\paragraph{\papertext{Quantitative analysis.}{定量分析。}}
\papertext{Table~\ref{tab:mmad} demonstrates ADClaw's broad capabilities in anomaly understanding on MMAD. ADClaw achieves the highest average accuracy and ranks among the top two on all seven tasks, leading in anomaly discrimination, defect localization, defect description, defect analysis, and object analysis. Compared with the same Qwen3.5-9B backbone, the gains are particularly pronounced in defect classification and localization, suggesting that hypothesis-specific checks help connect visual deviations to concrete defect types and spatial locations. Normal references provide a basis for judging deviations, while programmable inspection supplies targeted observations, together supporting anomaly understanding from recognition to defect description and analysis.}{表~\ref{tab:mmad} 展示了 ADClaw 在 MMAD 上较为全面的异常理解能力。ADClaw 取得了最高平均准确率，并在全部七项任务中位列前二，其中异常判别、缺陷定位、缺陷描述、缺陷分析及对象分析五项达到最优。相较于相同的 Qwen3.5-9B 主干，缺陷分类与定位的提升尤为明显，表明针对异常假设构造检查，有助于将视觉差异与具体缺陷类型及空间位置联系起来。正常参考提供判断偏差的依据，可编程检查进一步补充针对性观测，二者共同支持从异常识别到缺陷描述与分析的多层次理解。}

\vspace{-0.3em}
\paragraph{\papertext{Qualitative analysis.}{定性分析。}}
\papertext{Figure~\ref{fig:mmad_case} traces how executable checks revise an initial missing-screw hypothesis. The agent combines a retained full-image anomaly response with part masks to examine the suspected components, then checks composition and component integrity. Evidence that all expected components are present rejects the initial hypothesis, while the washer's localized response and shape comparison support a broken-washer diagnosis. The trace illustrates how the workbench reuses intermediate results and HER directs a subsequent check toward the unresolved defect before the final answer is produced.}{图~\ref{fig:mmad_case} 展示可执行检查如何修正初始螺钉缺失假设。智能体将已保存的全图异常响应与部件掩码结合，检查疑似部件，并进一步核验组成和部件完整性。所有预期部件均存在的证据否定了初始假设，而垫圈的局部异常响应与形状比较支持垫圈破损的诊断。该轨迹展示检测工作台如何复用中间结果，以及 HER 如何在最终作答前引导针对未决缺陷的后续检查。}

\begin{table}[t]
    \centering
    \setlength{\abovecaptionskip}{0pt}
    \caption{
        \papertext{\textbf{Inspection strategies on MVTec LOCO AD.} B adds predefined inspection compositions to A. All settings share visual tools and evidence.}{\textbf{MVTec LOCO AD 上的检查策略。}B 在 A 的基础上增加预定义检查操作组合。各设置共享视觉工具与证据。}
    }
    \label{tab:programmable_inspection}
    \footnotesize
    \setlength{\tabcolsep}{2pt}
    \renewcommand{\arraystretch}{1.15}

    \begin{tabularx}{\columnwidth}{Xcccc}
        \toprule
        \multirow{2}{*}{\textbf{\papertext{Strategy}{策略}}}
        & \multicolumn{3}{c}{\textbf{I-AUROC (\%, $\uparrow$)}}
        & \multirow{2}{*}{\papertext{\shortstack{\textbf{Calls}\\\textbf{(mean/image)}}}{\shortstack{\textbf{调用次数}\\\textbf{（每张图像）}}}} \\
        \cmidrule(lr){2-4}
        & \papertext{Overall}{总体} & \papertext{Logical}{逻辑异常} & \papertext{Structural}{结构异常} & \\
        \midrule
        A. \papertext{Adaptive}{自适应}
        & 80.91 & 78.36 & 83.64 & 5.03 \\
        B. \papertext{A + Comp.}{自适应＋组合}
        & 82.38 & 80.55 & 84.47 & 4.27 \\
        \midrule
        \rowcolor{black!6}
        C. \textbf{\papertext{ADClaw}{ADClaw}}
        & \textbf{84.10} & \textbf{83.01} & \textbf{85.21} & 4.31 \\
        \bottomrule
    \end{tabularx}
\end{table}

\begin{table}[t]
    \centering
    \setlength{\abovecaptionskip}{0pt}
    \caption{
        \papertext{\textbf{View-selection coverage under matched budgets.} Defect and Surface denote visible defect-region and input-scan surface coverage (\%).}{\textbf{相同视图预算下的覆盖率比较。}报告缺陷区域与输入扫描表面的可见覆盖率（\%）。}
    }
    \label{tab:geometry_view_coverage}
    \footnotesize
    \setlength{\tabcolsep}{2pt}
    \renewcommand{\arraystretch}{1.15}

    \begin{tabularx}{\columnwidth}{Xcccc}
        \toprule
        \multirow{2}{*}{\textbf{\papertext{Strategy}{策略}}}
        & \multicolumn{3}{c}{\textbf{\papertext{Defect}{缺陷} ($\uparrow$)}}
        & \textbf{\papertext{Surface}{表面} ($\uparrow$)} \\
        \cmidrule(lr){2-4}
        \cmidrule(lr){5-5}
        & \papertext{2 Views}{2 个视图} & \papertext{4 Views}{4 个视图} & \papertext{6 Views}{6 个视图} & \papertext{6 Views}{6 个视图} \\
        \midrule
        A. \papertext{Uniform}{均匀采样}
        & 56.80 & 74.10 & 83.50 & 89.20 \\
        B. \papertext{Coverage}{覆盖优先}
        & 62.40 & 80.20 & 88.80 & \textbf{94.10} \\
        \midrule
        \rowcolor{black!6}
        C. \textbf{\papertext{Geometry}{几何引导}}
        & \textbf{71.30} & \textbf{87.60} & \textbf{92.40} & 91.80 \\
        \bottomrule
    \end{tabularx}
\end{table}

\begin{table}[t]
    \centering
    \setlength{\abovecaptionskip}{0pt}
    \caption{
        \papertext{\textbf{Inspection budgets and inference cost.} \mbox{ADClaw ($n$)} allows at most $n$ steps; Selective adds HER-guided continuation. Latency and steps are means. \textbf{Bold} marks highest completion or lowest cost.}{\textbf{检查预算与推理开销。}ADClaw（$n$）表示最多执行 $n$ 步；选择性续检由 HER 引导。延迟与步数均为均值。\textbf{粗体}标记最高完成率或最低开销。}
    }
    \vspace{0.2em}
    \label{tab:inspection_budget}
    \footnotesize
    \setlength{\tabcolsep}{2pt}
    \renewcommand{\arraystretch}{1.15}

    \begin{tabularx}{\columnwidth}{Xcccc}
        \toprule
        \textbf{\papertext{Setting}{设置}}
        & \papertext{\shortstack{\textbf{Completion}\\\textbf{(\%)}}}
                     {\shortstack{\textbf{完成率}\\\textbf{（\%）}}}
        & \papertext{\shortstack{\textbf{Latency}\\\textbf{(s)}}}
                     {\shortstack{\textbf{延迟}\\\textbf{（秒）}}}
        & \papertext{\shortstack{\textbf{Steps}\\\textbf{(mean)}}}
                     {\shortstack{\textbf{步数}\\\textbf{（平均）}}}
        & \papertext{\shortstack{\textbf{Plan. tokens}\\\textbf{(k, in/out)}}}
                     {\shortstack{\textbf{规划 token}\\\textbf{（千，入/出）}}} \\
        \midrule
        \papertext{Fixed Plan}{固定计划}
        & 91.33 & \textbf{21.70} & 6.00 & \textbf{3.44 / 0.15} \\
        \papertext{Adaptive}{自适应}
        & 92.00 & 23.50 & 6.00 & 8.46 / 0.22 \\
        \midrule
        \papertext{\mbox{ADClaw (4)}}{ADClaw（4）}
        & 92.00 & 25.40 & \textbf{3.68} & 11.10 / 0.53 \\
        \papertext{\mbox{ADClaw (8)}}{ADClaw（8）}
        & 92.67 & 29.80 & 5.00 & 14.60 / 0.74 \\
        \papertext{\mbox{ADClaw (12)}}{ADClaw（12）}
        & 93.33 & 31.20 & 5.24 & 15.30 / 0.78 \\
        \midrule
        \rowcolor{black!6}
        \textbf{\papertext{Selective}{选择性续检}}
        & \textbf{94.00} & 27.40 & 4.40 & 12.50 / 0.61 \\
        \bottomrule
    \end{tabularx}
\end{table}

\begin{figure}[t]
    \centering
    \includegraphics[width=\columnwidth]{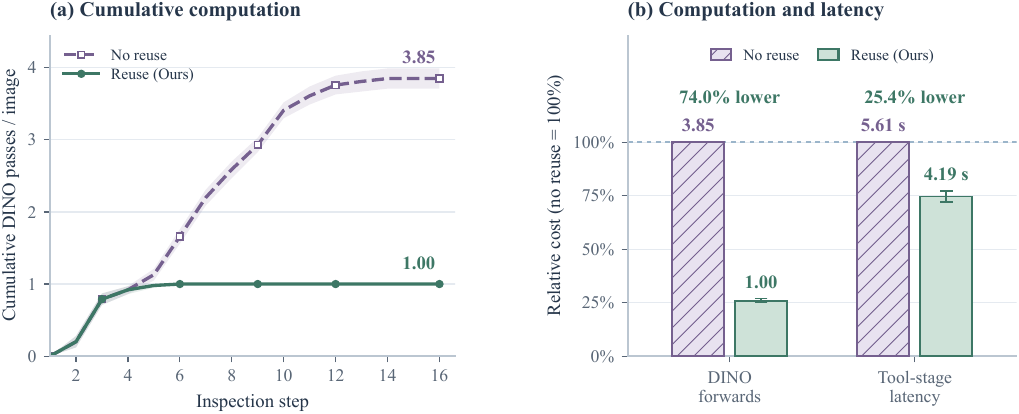}
    \caption{
        \papertext{\textbf{Efficiency of evidence reuse.}
        (a) Cumulative DINO passes. (b) Computation and tool-stage
        latency relative to no reuse. Raw means are annotated;
        bands and bars show image-level bootstrap 95\% confidence intervals.}{\textbf{证据复用的效率。}(a) 累积 DINO 前向计算次数。(b) 相对于不复用设置的计算量与工具阶段延迟。图中标注原始均值；阴影带与误差线表示通过图像级 bootstrap 估计的 95\% 置信区间。}
    }
    \label{fig:workbench_efficiency}
\end{figure}

\subsection{\papertext{Ablation Study}{消融实验}}
\label{sec:exp_ablation}

\paragraph{\papertext{Contribution of key mechanisms.}{关键机制的贡献。}}
\papertext{Table~\ref{tab:core_ablation} evaluates the Structured Normality Prior (SNP), the Inspection Workbench (IW), and Hypothesis-Guided Evidence Reasoning (HER). Removing SNP lowers Defect accuracy from 79.21\% to 75.85\%, supporting the role of normality priors in defect understanding. Removing IW reduces it to 76.65\%, consistent with the value of reusable evidence. Removing HER yields the largest Anomaly drop, from 74.98\% to 70.80\%. Object accuracy varies less across variants.}{表~\ref{tab:core_ablation} 评估结构化正常性先验（SNP）、检测工作台（IW）和假设引导的证据推理（HER）。移除 SNP 后，缺陷准确率由 79.21\% 降至 75.85\%，支持正常性先验在缺陷理解中的作用。移除 IW 使其降至 76.65\%，体现可复用证据的价值。移除 HER 导致最大异常判别降幅，由 74.98\% 降至 70.80\%。这些变体的对象准确率变化相对较小。}

\vspace{-0.5em}
\paragraph{\papertext{Effect of stateful evidence-driven inspection.}{有状态证据驱动检查的作用。}}
\papertext{Table~\ref{tab:inspection_strategy} compares inspection strategies with the same backbone, tools, normal references, preprocessing, and evaluation. Fixed Tool Pipeline executes a predefined sequence. Adaptive Tool Calling selects tools from earlier observations, treating their outputs as transient feedback. ADClaw retains structured evidence for reuse, hypothesis revision, and consistency validation. Adaptive selection increases the overall mean from 78.73\% to 80.32\%, and ADClaw raises it to 82.11\%. Relative to adaptive calling, ADClaw gains 2.58, 2.06, and 0.85 percentage points on Anomaly, Defect, and Object, respectively, illustrating the value of evidence organization for defect-oriented inspection.}{表~\ref{tab:inspection_strategy} 在相同主干、工具、正常参考、预处理及评估下比较检查策略。固定工具流程执行预定序列，自适应工具调用依据已有观测选择工具，将输出作为临时反馈。ADClaw 保留结构化证据，用于复用、假设修正和一致性验证。自适应选择将总平均由 78.73\% 提高至 80.32\%，ADClaw 进一步提高至 82.11\%。相较于自适应调用，ADClaw 在异常、缺陷及对象组分别提升 2.58、2.06 和 0.85 个百分点，支持证据组织对缺陷检查的价值。}

\vspace{-0.5em}
\paragraph{\papertext{Contribution of programmable inspection.}{可编程检查的贡献。}}
\papertext{Table~\ref{tab:programmable_inspection} examines inspection strategies with
the same visual tools and available evidence. Adding predefined tool compositions to adaptive calling
improves detection while reducing visual-tool calls, suggesting that organizing
inspection operations contributes beyond adaptive tool selection.
Programmable inspection further improves performance at a nearly identical
call count, with larger gains on logical than structural anomalies. This
pattern is consistent with the value of flexible composition for checking
relationships and constraints that individual tool outputs may not fully
capture. Overall, ADClaw improves I-AUROC over adaptive tool calling by
3.19 percentage points while using fewer visual-tool calls.}{表~\ref{tab:programmable_inspection} 在相同视觉工具与可用证据条件下比较不同检查策略。
在自适应工具调用基础上增加预定义检查操作组合，在减少视觉工具调用的同时改善了检测表现，表明检查操作的组织方式能够在自适应工具选择之外提供收益。
可编程检查在调用次数几乎相同的情况下进一步提升性能，且逻辑异常上的增益大于结构异常。
这一现象与灵活组合检查操作的作用相一致：其有助于检验单个工具输出可能无法充分反映的关系与约束。
总体而言，ADClaw 相较于自适应工具调用，以更少的视觉工具调用获得了 3.19 个百分点的整体 I-AUROC 提升。}

\vspace{-0.5em}
\paragraph{\papertext{Effect of geometry-guided view selection.}{几何引导视图选择的作用。}}
\papertext{Table~\ref{tab:geometry_view_coverage} compares view-selection strategies
under matched view budgets. Coverage is computed over the unique input-scan points jointly visible across the selected views; defect coverage applies the same calculation to annotated defect points. Uniform-angle sampling uses evenly spaced viewing directions; surface-coverage-first selection prioritizes additional visible surface, while geometry-guided selection targets suspected anomalous regions. Geometry-guided selection consistently achieves
the highest defect coverage, with its largest advantage under the two-view
budget, exceeding surface-coverage-first selection by 8.90 percentage points.
This suggests that geometric guidance is particularly useful when only a few
views are available. At six views, geometry-guided selection maintains broad input-scan surface coverage while providing greater defect visibility than surface-coverage-first selection. This pattern supports allocating views to geometrically suspicious regions in addition to expanding overall surface coverage. These results demonstrate that geometry-guided selection improves defect visibility within a limited view budget.}{表~\ref{tab:geometry_view_coverage} 在相同视图预算下比较不同视图选择策略。覆盖率根据所选视图联合可见的输入扫描点计算，重复可见点仅计一次；缺陷覆盖率以输入扫描中的标注缺陷点为统计对象。均匀角度采样按预设角度间隔选择观察方向，表面覆盖优先策略侧重新增可见表面，几何引导策略则优先观察疑似异常区域。
几何引导策略在各预算下均取得最高的缺陷覆盖率，且在两视图条件下优势最大，较表面覆盖优先策略提高 8.90 个百分点。
这表明，几何引导在可用视图较少时具有更明显的价值。
在六视图条件下，几何引导在保持较广输入扫描表面覆盖的同时，取得高于表面覆盖优先策略的缺陷可见覆盖率。
这一趋势支持在扩展整体表面覆盖的同时，将视图分配给几何上可疑的区域。
上述结果支持通过几何引导，在有限视图预算内提高缺陷的可见覆盖率。}

\vspace{-0.5em}
\paragraph{\papertext{Inspection budgets and selective continuation.}{检查预算与选择性继续检查。}}
\papertext{Table~\ref{tab:inspection_budget} compares the computational costs of different inspection strategies and budgets on 2,000 images sampled from MMAD. A workflow is considered complete when it executes successfully and produces a final report. Completion rates are averaged across repeated runs, while latency, executed steps, and planning-token consumption are averaged over all attempts. Fixed Plan follows a fixed initial plan, whereas Adaptive selects tools based on observation feedback. All ADClaw variants retain HER for evidence reasoning. Selective further uses the evidence state to determine whether additional inspection is needed, with the same 12-step limit as ADClaw (12). Within ADClaw, increasing the budget allows
more inspection and verification, improving workflow completion at the cost of
higher latency and planning-token consumption. This motivates adapting inspection
depth to the evidence requirements of each workflow. HER-guided selective continuation uses evidence readiness and completeness to decide whether to continue inspection. It prompts further verification when evidence is insufficient and ends inspection once the decision requirements are satisfied.
Compared with the 12-step configuration, it reduces mean latency by 12.2\% while
achieving a workflow completion rate of 94.00\%, together with lower planning-token
consumption and fewer executed steps. These results illustrate how HER coordinates inspection depth with evidence needs, enabling ADClaw to maintain a high completion rate while controlling average inference cost by avoiding unnecessary follow-up verification.}{表~\ref{tab:inspection_budget} 在从 MMAD 抽样的 2,000 张图像上比较不同检查策略与预算。完成率取多轮评测均值；延迟、步数与规划 token 对全部工作流尝试取均值。固定计划采用固定初始计划，自适应依据观测反馈选择工具；完成率表示成功执行并生成最终报告的比例。各 ADClaw 设置均保留 HER 推理，选择性续检在相同的 12 步上限内进一步依据证据状态控制是否继续检查。
在 ADClaw 中，增加预算为进一步检查与验证提供了空间，有助于提高流程完成率，
同时也带来了更高的延迟和规划 token 消耗。
这说明，根据各流程的证据需求调整检查深度，是协调完成率与计算开销的重要途径。
HER 引导的选择性继续检查依据当前证据是否足以支持决策、必要的证据检查是否完成，
判断是否需要进一步检查，从而在证据不足时支持补充验证，在满足决策要求后及时结束检查。
相较于最多 12 步的配置，该策略将平均延迟降低了 12.2\%，同时达到 94.00\% 的流程完成率，
并减少了规划 token 消耗与实际执行步数。
这些结果支持 HER 在协调检查深度与证据需求方面的作用，
使 ADClaw 能够在维持较高流程完成率的同时控制平均推理开销。}

\vspace{-0.5em}
\paragraph{\papertext{Efficiency of reusable evidence.}{可复用证据的效率。}}

\papertext{We measure evidence reuse through paired, fixed-trace replay on 1,000 MVTec LOCO
images. Both settings execute the same recorded actions and
differ only in reusing persisted full-image DINO responses for part-level
projections. In Figure~\ref{fig:workbench_efficiency}, reuse reduces mean DINO
forward passes from 3.85 to 1.00 per image (\textbf{74.0\%}) and tool-stage
replay latency from 5.61 to 4.19 seconds (\textbf{25.4\%}). Models and
normal references are preloaded; timing excludes Qwen planning
and subsequent postprocessing. All 1,000 paired runs pass output-equivalence
checks, preserving the recorded actions and outputs.
The workbench reduces repeated visual computation while preserving
evidence for subsequent inspection.}{我们通过对 1,000 张 MVTec LOCO 图像进行配对的固定轨迹回放，评估证据复用的效果。两种设置执行相同的记录动作，唯一区别是在部件级投影中是否复用已持久化的全图 DINO 响应。如图~\ref{fig:workbench_efficiency} 所示，复用将每张图像的平均 DINO 前向计算次数由 3.85 降至 1.00（减少 74.0\%），并将工具阶段的回放延迟由 5.61 秒降至 4.19 秒（降低 25.4\%）。模型与正常参考均预先加载；计时不包含 Qwen 规划和后续后处理。全部 1,000 组配对运行均通过输出等价性检查，保持了记录动作及输出的一致性。检查工作台在保留供后续检查使用的证据的同时，减少了重复视觉计算。}

\section{\papertext{Conclusion}{结论}}
\papertext{We presented ADClaw as an evidence-grounded framework for industrial anomaly detection and reasoning that writes and executes code to combine perception tools with prior observations into checks of anomaly hypotheses, and evaluated it across 2D, 3D, and anomaly-understanding benchmarks. The results show that programmable inspection and intermediate evidence validation can improve defect-oriented reasoning while maintaining competitive detection and localization performance, and that retaining and reusing inspection results can reduce redundant visual computation during multi-step inspection. By combining code-driven inspection with the retention and reuse of intermediate results, ADClaw offers an effective approach to tool-augmented industrial inspection and multi-stage diagnostic reasoning, particularly when diagnosis requires combining local measurements and checking part relationships and normality constraints. Future work will improve local defect inspection, cross-modal consistency, and generalization to unseen industrial categories and inspection settings.}{本文提出了以证据为依据的工业异常检测与推理框架 ADClaw，通过编写并执行代码，结合感知工具与已有观测，构造针对异常假设的检查，并在二维、三维及异常理解基准上进行了评估。实验结果表明，可编程检查与中间证据验证能够在保持有竞争力的检测与定位性能的同时，改善面向缺陷的推理；保留并复用检查结果还能够减少多步检查中的重复视觉计算。通过将代码驱动的检查与中间结果的保留和复用相结合，ADClaw 为工具增强的工业检查和多阶段诊断推理提供了一种有效途径，尤其适用于需要组合局部测量、检验部件关系与正常性约束的场景。未来工作将聚焦于更稳健的局部缺陷检查、更强的跨模态一致性验证，以及对未见工业类别和检查场景的泛化。}

\section*{\papertext{LLM Usage Statement}{大语言模型使用声明}}
\papertext{We shared the manuscript with OpenAI (ChatGPT/GPT) for reviewer-style feedback. For writing and editing, code and implementation, and experimental design, execution, and analysis, we used AI tools to edit or refine work we had produced (Assist). We used no LLM or AI tools for related work or citations; figures, diagrams, or other visual material; data preparation, including synthetic data, dataset creation, and annotation; or research ideas, hypotheses, claims, or conclusions. We take full responsibility for all submitted content.}{我们将稿件提供给 OpenAI（ChatGPT/GPT），以获取审稿式反馈。在文字撰写与编辑、代码与实现，以及实验设计、执行与结果分析方面，我们使用 AI 工具编辑或完善由作者完成的内容（Assist）。我们未在以下方面使用大语言模型或 AI 工具：相关工作与引用；图表与其他视觉材料；数据准备，包括合成数据、数据集构建与标注；以及研究想法、假设、论断或结论。我们对提交的全部内容承担完全责任。}

{
    \small
    \bibliographystyle{ieeenat_fullname}
    \bibliography{main}
}

\end{document}